\documentclass{article}

\usepackage{arxiv}

\usepackage[utf8]{inputenc} 
\usepackage[T1]{fontenc}    
\usepackage{hyperref}       
\usepackage{url}            
\usepackage{booktabs}       
\usepackage{amsfonts}       
\usepackage{nicefrac}       
\usepackage{microtype}      
\usepackage{lipsum}
\usepackage{times}
\usepackage{latexsym}
\usepackage{hyperref}
\usepackage{graphicx}
\usepackage{tabularx}
\usepackage{url}

\title{Claim Extraction in Biomedical Publications using Deep Discourse Model and Transfer Learning}

\author{
Titipat Achakulvisut  \\
  Department of Bioengineering \\ 
  University of Pennsylvania \\
  Philadelphia, PA, USA\\
  \texttt{titipata@seas.upenn.edu} \\
  \And
  Chandra Bhagavatula \\
  Allen Institute for Artificial Intelligence\\ 
  Seattle, WA, USA \\ 
  \texttt{chandrab@allenai.org}
  \AND
  Daniel E. Acuna \\ School of Information Studies\\ 
  Syracuse University \\
  Syracuse, NY, USA \\
  \texttt{deacuna@syr.edu} \\
  \And
  Konrad Kording \\ Department of Bioengineering\\ 
  University of Pennsylvania\\ 
  PA, Pennsylvania, USA \\
  \texttt{koerding@gmail.com}
}

\begin{document}
\maketitle

\begin{abstract}
Claims are a fundamental unit of scientific discourse. The exponential growth in the number of scientific publications makes automatic claim extraction an important problem for researchers who are overwhelmed by this information overload. Such an automated claim extraction system is useful for both manual and programmatic exploration of scientific knowledge. In this paper, we introduce a new dataset of 1,500 scientific abstracts from the biomedical domain with expert annotations for each sentence indicating whether the sentence presents a scientific claim. We introduce a new model for claim extraction and compare it to several baseline models including rule-based and deep learning techniques. Moreover, we show that using a transfer learning approach with a fine-tuning step allows us to improve performance from a large discourse-annotated dataset. Our final model increases F1-score by over 14 percent points compared to a baseline model without transfer learning. We release a publicly accessible tool for discourse and claims prediction along with an annotation tool. We discuss further applications beyond biomedical literature.
\end{abstract}


\section{Introduction}

Claims are a fundamental unit of scientific discourse. These days, researchers have a hard task keeping track of new scientific claims in an exponentially increasing number of publications. Automatic extraction of claims from scientific articles promises to alleviate some of these problems. Additionally, automated claim extraction could support automated knowledge exploration, efficient reading \cite{dernoncourt2017pubmed}, and text summarization \cite{teufel2002summarizing}. Such automatic tool might promise to augment decision making in science for readers, reviewers and funding \cite{weber2019applying}. The recent DARPA's Systematizing Confidence in Open Research and Evidence \href{https://www.darpa.mil/program/systematizing-confidence-in-open-research-and-evidence}{(SCORE) program} highlights that this task is important beyond scientific research.

Claims can be thought of as being part of the scientific discourse. Discourse usually appears as a sequence of sentences or texts; researchers use this sequential information to analyze discourse \cite{webber2012discourse}. Many studies have proposed a variety of schemes to classify scientific discourses \cite{liakata2010corpora, teufel2009towards}. Multiple natural language features can be used, ranging from a sentence to a phrase to an entity. In a sense, claim extraction can be considered a similar task. 

Automated methods for classifying scientific discourse heavily rely on data. Most proposed discourse datasets are limited in their size and tools \cite{neves2019evaluation}. This is because the annotation task is challenging for general readers which makes the production of such datasets laborious. High-quality datasets, moreover, can only be created by domain experts or readers with scientific background. These datasets, therefore, are frequently focused on a narrow domain. For example, previous annotated domains include physical chemistry \cite{liakata2010corpora}, computer science \cite{liakata2010corpora}, computer graphics \cite{fisas2015discoursive}, engineering \cite{dayrell2012rhetorical}, life science, and biomedicine. It would be beneficial to produce larger datasets with more standard tooling to explore more sophisticated automated claim extraction methods with machine learning and deep learning. 

In this paper, we introduce a scientific claim extraction method and a dataset of 1,500 scientific papers with expertly annotated claims. This is the largest claim-annotated dataset for biomedical documents that we know of. We build on a neural discourse tagging model based on a pre-trained Bidirectional LSTM (Bi-LSTM) network with Conditional Random Field (CRF) \cite{chen2017improving} and transfer the representations to a claim extraction model. Our fine-tuned model achieves 47\% higher F1-score compared to the rule-based method presented in the previous research \cite{sateli2015semantic}. We show that transfer learning helps performance significantly compared to a similar architecture without transfer. We make the code for claim extraction, our dataset, and our annotation tool publicly available to the community \footnote{\texttt{\href{https://github.com/titipata/detecting-scientific-claim}{https://github.com/titipata/detecting-scientific-claim}}}. We discuss how to extend this framework to work with larger bodies of texts and other domains.


\begin{figure}
\centering
\includegraphics[width=0.5\linewidth]{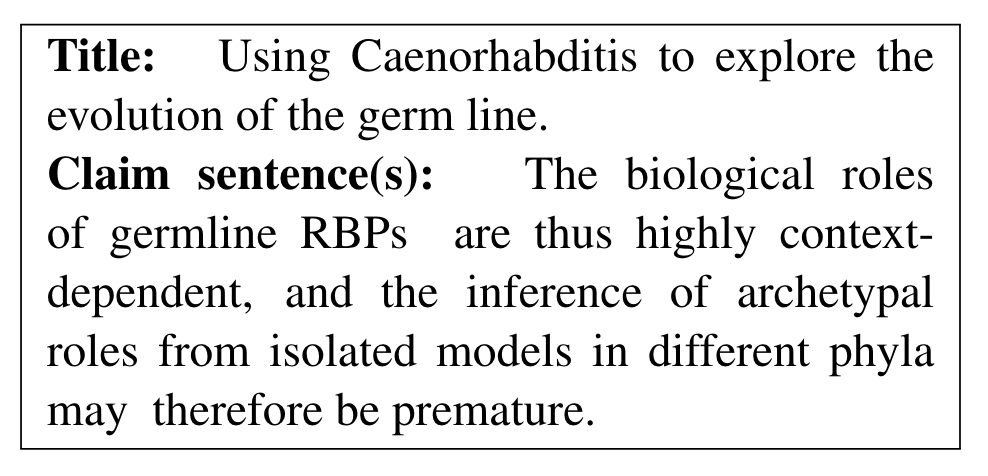}
\caption{Example of annotated claims for a paper in our dataset.}
\label{fig:example_claim}
\end{figure}

\section{Related Work}

Previous research has attempted to extract claims and premises from news \cite{Habernal2014ArgumentationMO, Sardianos2015ArgumentEF}, social media \cite{Dusmanu2017ArgumentMO}, persuasive essays and scientific articles \cite{stab2014argumentation}, and Wikipedia \cite{freard2010role, Thorne2018FEVERAL}. This task is generally called \textit{Argumentation Mining}. However, in the biomedical domain, the emphasis and data availability have been substantially less prominent compared to these other domains. In particular, the claim extraction task has been dominated by rule-based techniques \cite{sateli2015semantic, yuan2019hclaime} and classic machine learning techniques \cite{liakata2010corpora, Guo2011AWA} whereas modern Argumentation Mining uses deep learning techniques (e.g., \cite{dehghani2017learning, Ratner2018TrainingCM, augenstein2018multi}). These more modern techniques have shown, for example, that weak supervision and transfer learning can help improve several text prediction tasks. It is unclear whether these improvements would translate into out specific task.

Recently, Thorne et al. \cite{Thorne2018FEVERAL} released the FEVER dataset that consists of factual claims from Wikipedia validated by crowd-workers. However, annotating scientific claims requires significant domain expertise rather than crowd-workers. Dernoncourt et al. \cite{dernoncourt2017pubmed} released \textsc{PubMedRCT}, a related dataset of scientific discourse where sentences are labeled as belonging to \textit{background}, \textit{introduction}, \textit{method}, \textit{result}, and \textit{conclusion} sections. However, this dataset is not expertly annotated. For claim extraction, then, we have limited expertly annotated datasets (see \cite{dasigi2017experiment, sateli2015semantic, liakata2010corpora}).

Several techniques have been proposed for causal claim extraction. In \cite{yuan2019hclaime}, the authors used rule-based claim extraction techniques and in \cite{yu2019detecting} they explored multiple deep learning architectures. In work outside of science,  \cite{li2019causality} proposed a Bi-LSTM deep learning architecture using triplets to extract causal claims. Deep learning techniques are known to require large amounts of data to work well, and therefore causal claim extract needs datasets of scale.

The scale of datasets directly related to claim extraction are comparatively small and domain limited.  \cite{liakata2010corpora} produced the CoreSC dataset containing 265 articles in physical chemistry and biochemistry. \cite{fisas2015discoursive} produced the Dr. Inventor dataset, which contains annotations of 40 computer graphics articles. \cite{dasigi2017experiment} introduced a dataset of 75 articles for predicting discourse using articles from PubMed. \cite{sateli2015semantic} presented a dataset with annotations for claims and contributions using full text of 30 articles from the computer science domains. Claim extraction for biomedical literature might benefit from these previous datasets.

There are several limitations in the past literature. First, most previous studies use datasets that are relatively small, ranging from 40 to a couple of hundreds annotated publications. Since this task is complex, experts are often needed to provide these annotations. This exacerbates the problem of small data sizes. This has led to fragmentation of the datasets whereas they are usually limited to specific domains such as computer science, computational linguistic, and chemistry. To make this important task have better automated techniques, we need to significantly increase the number of annotated instances, the quality of the annotation, and the sophistication of the techniques used. In this article, we explore these avenues of improvements.

\section{Discourse and claim extraction tasks}


In this section, we introduce the discourse prediction task and the claim extraction task. We now define some notation used throughout the article. Formally, an abstract is represented as a sequence of sentences $\mathbf{S} = \{S_1, ..., S_n\}$. In \textsc{PubMedRCT} dataset (used for pre-training), each sentence is associated with a discourse type $\{D_1, ..., D_k\}$, where $D_i \in $ \{\textit{Objective}, \textit{Introduction}, \textit{Methods}, \textit{Results}, or \textit{Conclusions}\}. The \textbf{discourse prediction task} is to predict the discourse types for a sequence of sentences from a new abstract. The \textbf{claim extraction task} is to predict whether a sentence $S_i$ is a claim (binary classification). As an example, figure \ref{fig:annotation} has an annotated abstract, which is a sequence of sentences, and with one or more sentences labeled as claims.

\section{A new expertly-annotated dataset of biomedical claims}

As part of the new claim extraction task, we introduce a novel dataset of expertly annotated claims in biomedical paper abstracts. While there are multiple definitions of scientific claims proposed in previous literature, we follow the definition of \cite{sateli2015semantic} to characterize a claim. This is, a claim is a statement that either (1) declares something is better, (2) proposes something new, or (3) describes a new finding or a new cause-effect relationship. One abstract can have multiple claims.

\subsection{An annotation tool for claims}

\begin{figure*}[h!]
\centering
\includegraphics[width=0.9\linewidth]{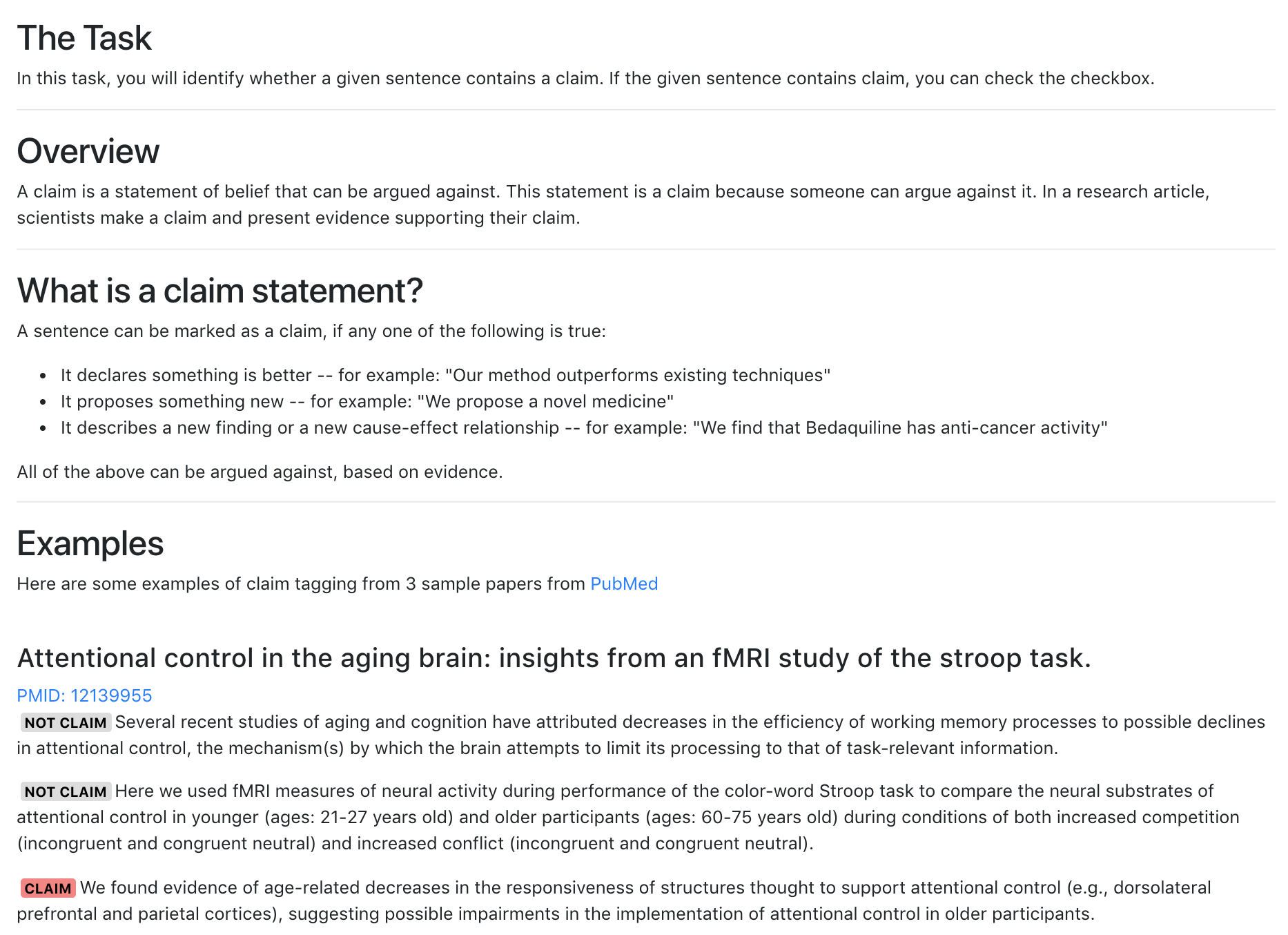}
\caption{{\bf Screenshot of the instruction of annotation tool}. The annotators are presented by the task instruction, definition of a claim, and examples of annotated documents before the task.}
\label{fig:task}
\end{figure*}

\begin{figure*}[ht!]
\centering
\includegraphics[width=0.9\linewidth]{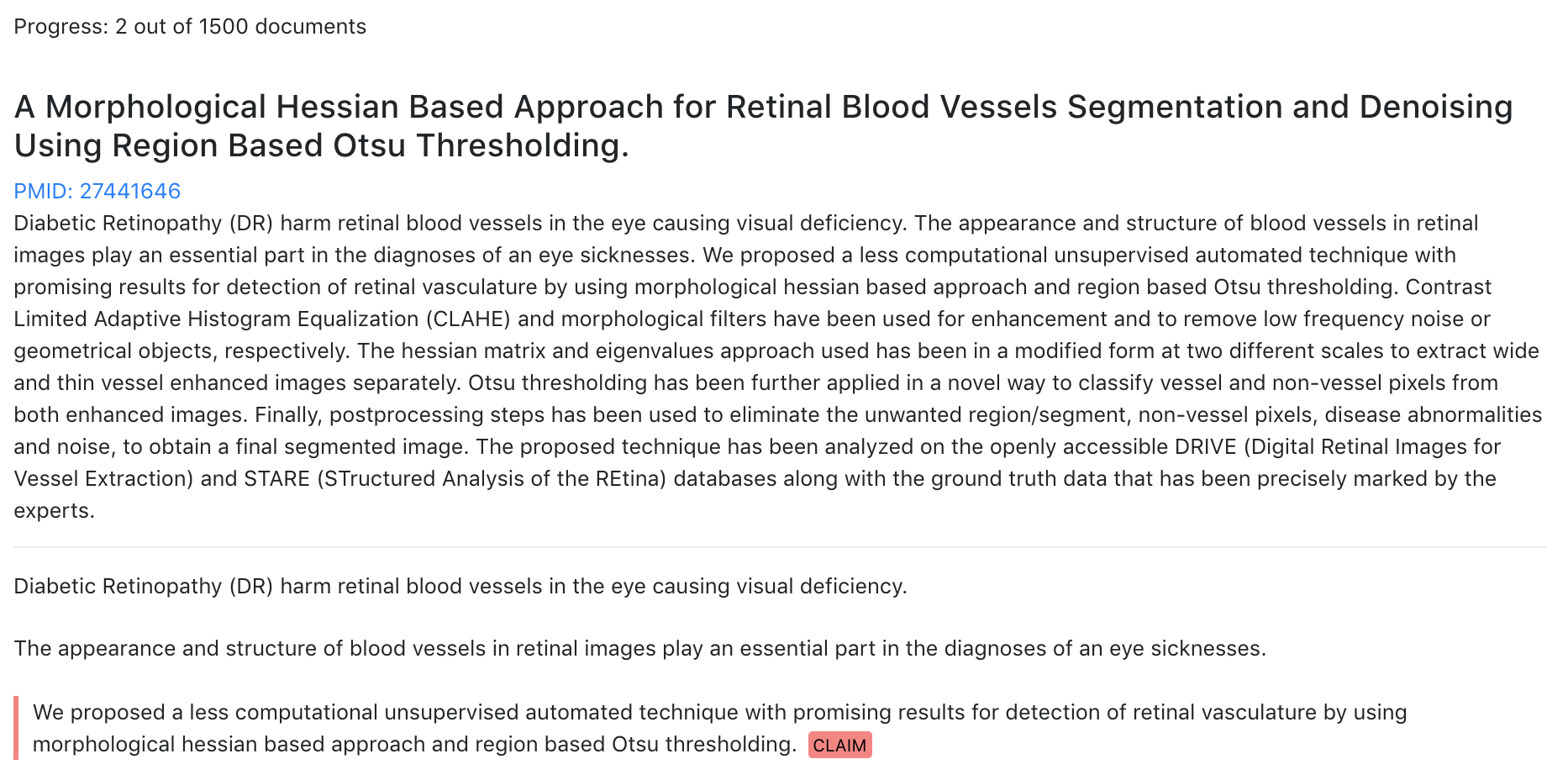}
\caption{{\bf Screenshot of the annotation tool}. The abstracts are sampled from MEDLINE database, sentences are pre-split, and annotators can select which sentences are claims.}
\label{fig:annotation}
\end{figure*}

The annotation processes required a special annotation tool that we now describe. The first page of the annotation tool contains the description of the task and examples. The second page contains title and abstract of a sample publication to be annotated. The abstract is pre-split into sentences using \texttt{spaCy} \cite{spacy2}. A screenshot of the annotation tool's instruction and annotation example is shown in figures \ref{fig:task} and \ref{fig:annotation}, respectively.

\subsection{Annotation results}

Three annotators with biomedical domain expertise and fluency in English are selected for the task. The annotators were presented with instructions and a few examples before the start. The abstracts are sampled from the top 110 biomedical venues sorted by a number of abstracts in the \href{https://www.nlm.nih.gov/bsd/medline.html}{MEDLINE database} from the year 2008 to 2018 parsed using the \textit{Pubmed Parser} library \cite{achakulvisut2015}. The dataset consists of 1,500 abstracts containing 11,702 sentences. 

\begin{figure*}[h!]
\centering
\includegraphics[width=0.5\linewidth]{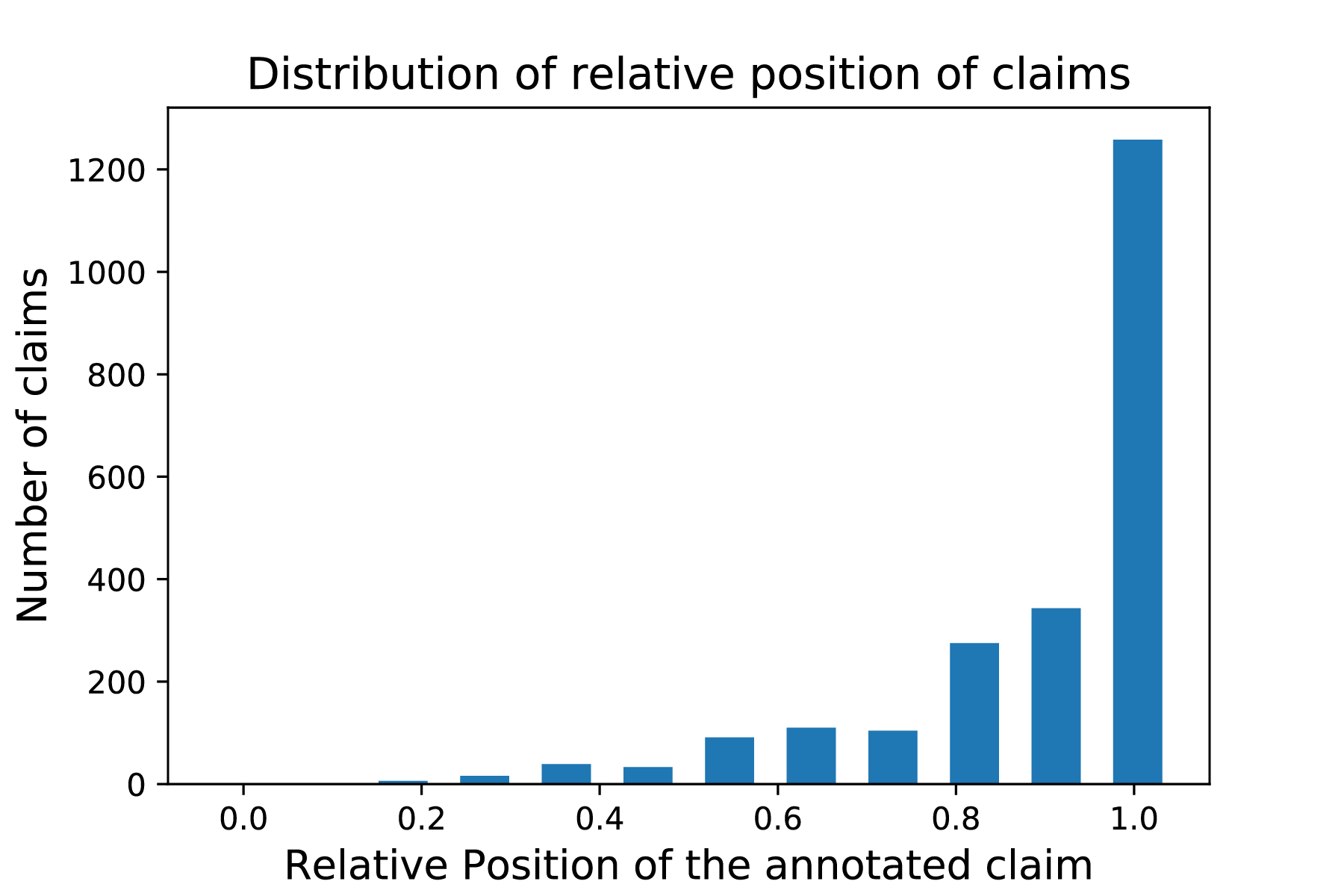}
\caption{{\bf Distribution of relative position of annotated claims in the dataset}. Around 55.3 percent of the claims are located in the last sentence and the rest are elsewhere in the abstract.}
\label{fig:claim_position}
\end{figure*}

We analyze the reliability of our annotators in several ways. The pairwise inter-annotator agreement (IAA) scores between the three annotators (Cohen's Kappa, $\kappa$, \cite{Carletta1996AssessingAO, artstein2008inter}) are 0.630, 0.575, and 0.678 respectively. The Fleiss's Kappa between all annotators is 0.629. The final label for training the claim prediction model is computed as the majority vote between all three annotators, producing a total of 2,276 claim sentences. This is a relatively low IAA score because of the nature of the task \cite{lauscher2018argument}. The distribution of the relative position of the claims identified through this process is shown in figure \ref{fig:claim_position}. Around 55.3\% of the claims are located in the last sentence. As a control, crowdsourced annotators with \textbf{no} biomedical background have a significantly lower $\kappa$ of 0.096 compared to experts. This shows that a great deal of background knowledge is required for the annotation. Thus, annotating abstract could be expensive because of this specialization.

\section{A new claim extraction model using transfer learning with fine-tuning from discourse prediction model}

Our new claim extraction model requires a pre-trained step and fine-tuning. Below, we explain both components.

\subsection{Discourse Prediction Model based on Structured Abstracts}


In the discourse prediction task, we use a sequence of the structured abstracts as an input $\mathbf{S}$ and try to predict a sequence of discourse outputs which can be one from any of the five classes: \textit{Objective, Introduction, Methods, Discussions}, or \textit{Conclusions}. One of the advantages of using this scheme is that there is a large corpus of structured abstracts from MEDLINE dataset. We want to see if we can exploit the overall structure of the abstracts by pre-training the model to predict the discourse using structured abstracts. A dataset containing this information has been released as \textsc{PubMedRCT} \cite{dernoncourt2017pubmed}. We use this dataset here.

To train a discourse prediction model, we use the \textsc{PubMedRCT} dataset \cite{dernoncourt2017pubmed} which contains discourse types associated with each sentence in 200,000 abstracts. Sentence and word tokenization are done using \texttt{spaCy} \cite{spacy2}. We experiment with two main neural architectures -- (i) vanilla Bi-LSTM model and (ii) Bi-LSTM with CRF layer (Bi-LSTM CRF) presented by \cite{chen2017improving}. The CRF layer takes the outputs at each timestep of the Bi-LSTM layer and uses them to jointly infer the most probable discourse type for all sentences in a given sequence of sentences. Transfer learning and fine tuning are applied to pre-trained discourse models to train the claim extraction model later on.

More specifically, the model consists of a Bi-LSTM network that predicts the class of the sentences. It takes in a sequence of words, transforms them into embeddings, and use them to produce forward and backward LSTM states. These states are passed through a feedforward network that predicts the class. When multiple sentences are combined, the class distributions per sentence are used by a Conditional Random Field (CRF) layer that predicts the most likely sequence of classes. This layer allows, for example, to learn that it is more likely to have a conclusion sentence after a result sentence---and not the other way around. This is a standanrd model used in several Natural Language Processing (NLP) tasks, especially entity recognition \cite{chen2017improving}.

\begin{figure}
\centering
\includegraphics[width=0.5\linewidth]{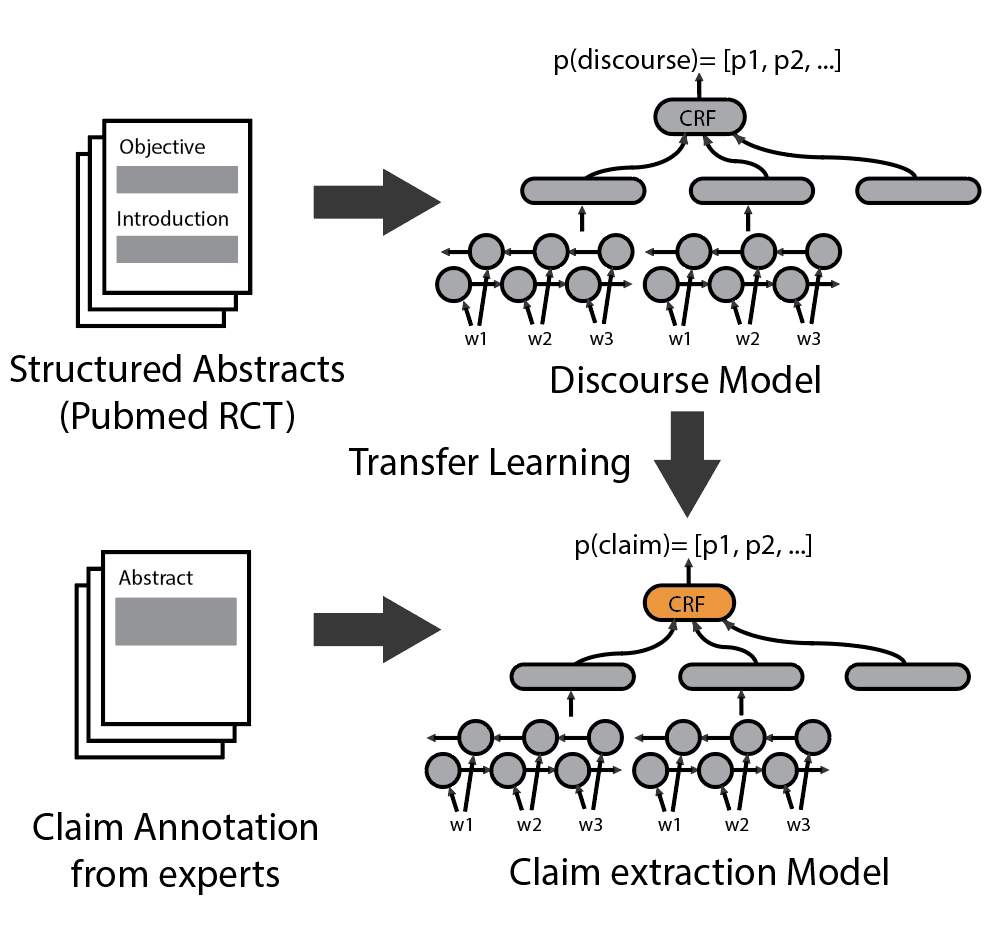}
\caption{{\bf Schematic of Transfer Learning with fine-tuning technique}. Here, we show an example schematic of transfer learning using Bi-LSTM CRF model. In discourse model, we use Bi-LSTM layers as representations of sentences. These representations are then passed through dense neural network layers with conditional random field layers and softmax activation functions to predict the discourse output probabilities. In transfer learning and fine-tuning, the last layer of the discourse classification model is replaced by the claim extraction model which predicts a binary output instead. Transfer learning and fine-tuning are applied to adapt from the learned discourse classification structure.}
\label{fig:schematic}
\end{figure}

\subsection{Claim extraction model using transfer learning from the discourse prediction model}

We propose transfer learning to train a claim extraction model from a pre-trained discourse model. In a few words, the discourse Bi-LSTM and Bi-LSTM CRF models are trained based on a \textsc{PubMedRCT} dataset using pre-trained GloVe word embeddings \cite{pennington2014glove} with 300 dimensions. We also experimented with 200-dimensional word embeddings training on PubMed \cite{moen2013distributional}. We train the discourse model with batch size of 64 with a reduce-on-plateau learning rate scheduler with factor of 0.5 and Adam optimizer \cite{kingma2014adam} with the learning rate of 0.001 for all experiments. All models are implemented using the AllenNLP library \cite{gardner2018allennlp}. After this training, we apply transfer learning and fine tuning to train the models on our expertly annotated dataset. Transfer learning is done by replacing the last layer of the pre-trained model (Fig. \ref{fig:schematic}) with the new layer to predict claim instead. We train the model while freezing parameters of the pre-trained model until it reaches lowest validation cross entropy loss. Then, we unfreeze all the layers to fine tune all parameters until loss updates converge. The schematic of the training process can be found in figure \ref{fig:schematic}.

\begin{table*}[ht!]
\centering
\small
\begin{tabular}{cc}
\begin{tabular}{|c|ccc|ccc|}
\hline
 & \multicolumn{3}{|c|}{ {\bf Validation}} & \multicolumn{3}{|c|}{ {\bf Test}} \\
{\bf Model} & {\bf Precision} & {\bf Recall} & {\bf F1-score} & {\bf Precision} & {\bf Recall} & {\bf F1-score} \\
\hline
Rule-based (Sateli et. al.) & 0.349 & 0.364 & 0.356 & 0.315 & 0.322 & 0.319 \\
Last sentence as a claim & 0.845 & 0.542 & 0.660 & 0.835 & 0.548 & 0.662\\
Sent Embedding & 0.605 & 0.641 & 0.623 & 0.624 & 0.674 & 0.648 \\ 
Sent Embedding + Discourse & 0.709 & 0.723 & 0.716 & 0.715 & 0.711 & 0.713 \\ 
Bi-LSTM CRF only annotation data & 0.778 & 0.521 & 0.624 & 0.701 & 0.609 & 0.652 \\
Bi-LSTM CRF Conclusion as Claim & 0.616 & 0.773 & 0.685 & 0.582 & 0.792 & 0.671 \\
Transfer Learning (PubMed) & 0.735 & 0.723 & 0.729 & 0.723 & \textbf{0.730} & 0.727 \\
Transfer Learning (Glove) & 0.738 & \textbf{0.765} & 0.751 & 0.762 & 0.729 & 0.745 \\ 
Transfer Learning CRF (PubMed) & 0.840 & 0.764 & 0.800 & \textbf{0.887} & 0.685 & 0.773 \\
Transfer Learning CRF (Glove) & \textbf{0.859} & 0.750 & \textbf{0.801} & 0.866 & 0.727 & \textbf{0.790}  \\
\hline
\end{tabular}
\end{tabular}
\caption{ {\bf Model Evaluation on claim annotation dataset.} We report the performance of the baseline rule-based model, sentence embedding model, Bi-LSTM CRF trained only on annotation dataset, sentence based transfer learning model, and our final transfer learning CRF model. The transfer learning model is pre-trained using one sentence as an input and fine-tuned it to our annotated claim dataset. Transfer Learning with CRF performance is pre-trained using a sequence of sentences from an abstract. Then, the discourse CRF model is fine-tuned to our annotated claim dataset.}
\label{tab:claim}
\end{table*}

\section{Experimental setup}

\subsection{Baseline models}

\textbf{Rule-based claim extraction (Rule-based in Table \ref{tab:claim})} We implement a baseline model using the rule-based claim extraction algorithm presented by Sateli et al. \cite{sateli2015semantic}.  It processes part-of-speech patterns and pre-defines a set of keywords that signal claim statements. We re-implement the algorithm in Python with the \texttt{spaCy} library \cite{spacy2}. This achieves a similar F1-score to that reported in the original paper (Table \ref{tab:claim}).

\textbf{Sentence embedding with discourse probability (Sent. Embedding + Discoursein Table \ref{tab:claim})} We implement another baseline model using the sentence classification technique presented by Arora et al. \cite{arora2016simple}. The sentence embedding is calculated by a weighted combination of the inverse word frequency in MEDLINE abstracts and word embeddings. The first principal component vector calculated from the embedding vectors of the training is subtracted to remove the shared common principal components. This removal has been shown to achieve better performance in sentence classification problems. To improve the classification performance, we concatenate sentence embeddings with discourse probabilities calculated by the discourse prediction model. Finally, we apply regularized logistic regression to predict the probability of a claim.

\textbf{Transfer learning without conditional random field layer (Transfer Learning in \ref{tab:claim})} To observe the improvement brought by the CRF layer in Bi-LSTM CRF model, we used a sentence based pre-trained model as an additional baseline model. The pre-trained discourse model is trained on a single sentence input instead of a sequence of sentences from an abstract. We then apply a transfer learning and fine tuning to train on our annotated claim dataset.

\subsection{Model Selection and Evaluation}

To evaluate the performance of the classifier, we perform experiments using 50\% of the corpus for training (750 articles), 25\% for validation (375 articles), and 25\% for testing (375 articles). We train and evaluate our propose model and baseline models on training and validation set using precision, recall, and F1-score. Since the class is highly imbalance, these metrics are used in previous articles to measure the performance of claim classifiers instead of accuracy score \cite{li2019causality}. Finally, we report the precision, recall, and F1-score of the final performance of the models on testing set (Table \ref{tab:claim}).

\subsection{Results}

In this work, we are trying to predict which sentences contain a claim. We have developed a new dataset and models, which we now test. Specifically, we compare the performance of a rule-based model, predicting last sentence as a claim model, sentence embedding model, sentence embedding with a concatenation of discourse probabilities (Sentence Embedding + Discourse) model, Bi-LSTM with CRF layer trained only using the claim extraction dataset model, Bi-LSTM with CRF layer where we use the conclusion class as a true label for claim model, transfer learning of vanilla Bi-LSTM of discourse prediction model, and transfer learning from Bi-LSTIM with CRF layers model. 

The rule-based model presented by Sateli et al. \cite{sateli2015semantic} achieves F1-score of 0.35. Compared to other models, it has substantially lower score (Table \ref{tab:claim}). This suggests that the rule-based approach built for the CS domain does not transfer well to the biomedical domain. Predicting last sentence as a claim gives a high precision but relatively low recall. We observe that the transfer learning with fine-tuning of pre-trained Bi-LSTM CRF model from discourse task achieves the best F1-score. The GloVe pre-trained word vector gives a slightly better performance compared to the PubMed-based vectors. These results suggest that overall discourse information is relevant and improves performance substantially. The difference between the F1-score of the best model without discourse and model with pre-trained discourse is 14\%.

\begin{figure*}[b!]
\centering
\includegraphics[width=0.9\linewidth]{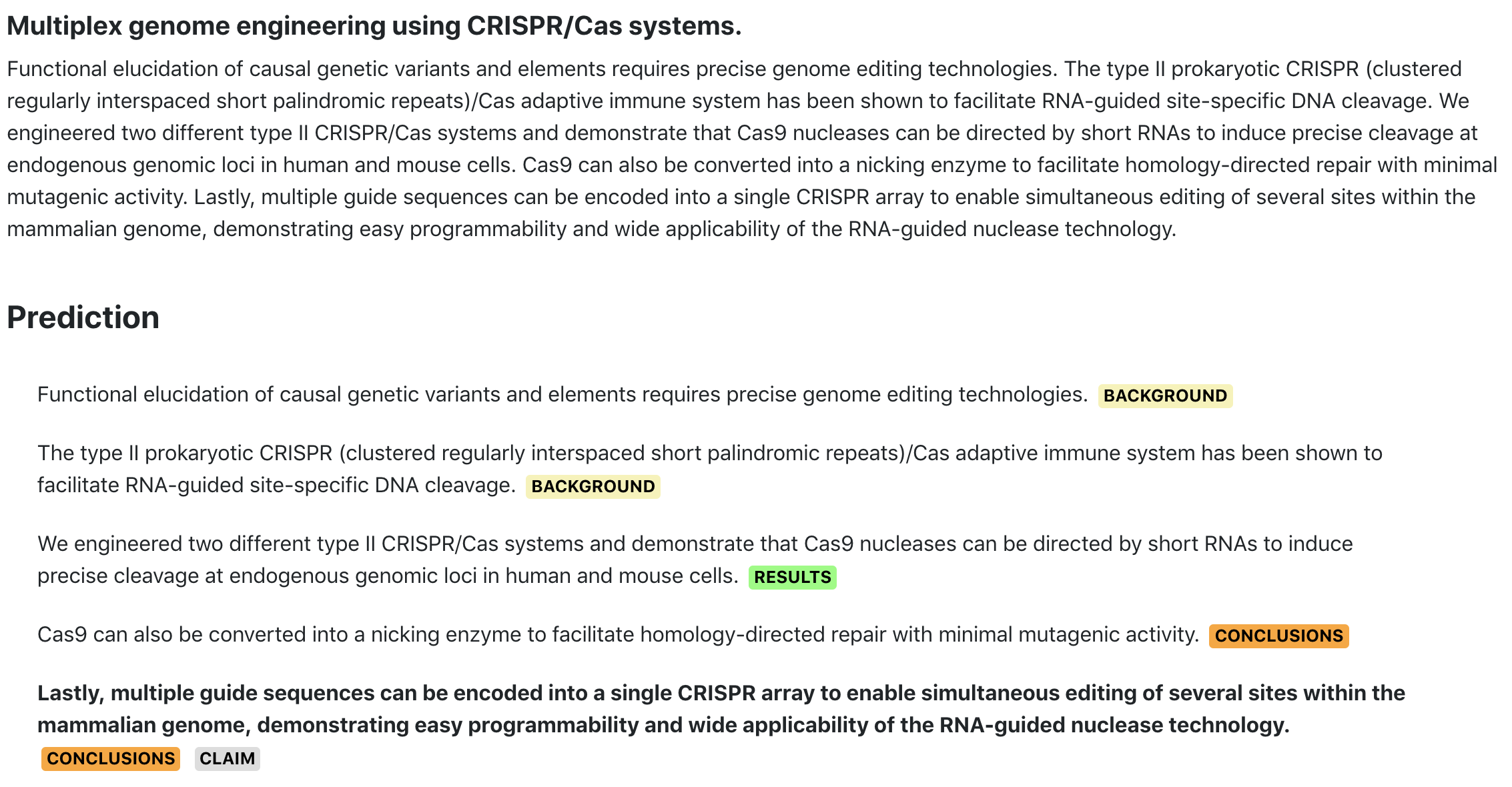}
\caption{{\bf Screenshot of the output from Discourse and Claim Prediction tool}. A sample discourse and claim prediction from CRISPR/Cas article published in \textit{Science} (PMID: \href{https://www.ncbi.nlm.nih.gov/pubmed/23287718}{23287718}) using transfer learning of Bi-LSTM CRF model with fine-tuning.}
\label{fig:screenshot}
\end{figure*}

Finally, we developed a web-based tool to perform the claim extraction based on the title and abstract of a publication. We show example screenshots of the output from claim prediction tool in figure \ref{fig:screenshot}. 

\textbf{Error Analysis} The best model correctly predicts all claims in abstracts for 60 percent of test articles. 31 percent of the articles has only one missclassified sentence. We now examine the rest of the cases where we misclassified more than one sentence. We found two types of errors. In one type of error, the model is unusually confident that the last sentence contains a claim. For example, for the article PMID: \href{https://www.ncbi.nlm.nih.gov/pubmed/25717311}{25717311} the model wrongly predicts the sentence \textit{"Results are discussed in the context of developmental theories of visual processing and brain-based research on the role of visual skills in learning to read."} as a claim when in reality it is just a result statement. In another type of error, the model predicts with low probability that claims are in the middle of the abstract. For example, for the abstract PMID: \href{https://www.ncbi.nlm.nih.gov/pubmed/24806295}{24806295}, it misses predicting the following sentence as a claim \textit{"... Our results also show that aMCI forms a boundary between normal aging and AD and represents a functional continuum between healthy aging and the earliest signs of dementia.".} even though it predicts the previous sentence as a claim correctly. Investigating the cause of this pattern and making modifications to address these issues are an area of future work.

\section{Discussion}

In this paper, we propose a framework for automated claim extraction in biomedical publications. As part of the task, we produce a new expertly annotated claim dataset for biomedical publications. We also create an annotation tool and a prediction tool for the model. We show that the pre-trained model from structured abstracts with transfer learning can achieve the best predictions. We also observe that learning the sequential nature of discourse, exploited by the conditional random field layer, can improve predictions further.

\subsection{Limitations of our work}

Our application is still limited to the abstract of a publication. Expanding our prediction to full text could be challenging due to the writing structure of a full publication. The claim can be anywhere in the article which makes the prediction harder due to large number of possible locations compared to an abstract. For example, the scientific writing in methods and discussion sections are different from each other. We speculate that the model will try to predict no sentences as claims since most sentences are not claims. The fine-tuning of the model to different parts of the publication might not generalize well. More data would be needed for this case.

Our dataset is still not as big as other standard NLP tasks such as co-reference resolution, entity extraction, and part-of-speech tagging. Since the annotation of biomedical publications requires background knowledge and a great deal of training, our results are a first step in solving these issues. We try to mitigate the difficulty of the task by creating user-friendly tools that can be easily used by annotators. However, it still takes approximately 50 hours per annotator to finish 1,500 abstracts. The agreement between annotators is not high (e.g., see \cite{artstein2008inter}) but still much higher than our control with crowd-sourced workers. Previous publications also reported similar low agreements \cite{fisas2015discoursive}. Anecdotally, we observed that we can improve labeling by giving annotators better instructions and more annotated examples.

There are several improvements that we foresee being addressed in the future. Due to idiosyncrasies of the dataset, one of the problems is that our best model is slightly biased towards predicting the last few sentences of an abstract as claims. While this is mostly correct for biomedical papers, it might not be the structure of other fields, such as Computer Science or Social Science. Also, we are not certain about why the word vectors learned from PubMed are no better than the GloVe vectors. We speculate that specializing the distributed representation of vectors to the domains should help more. We believe that this drop in performance can come from (i) the learning of the vectors---i.e. PubMed vectors use word2vec skip-gram model to capture information from surrounding words and GloVe vectors use a neural network to capture co-occurrence structure of words. The information about word co-occurrence might be more meaningful to capture the claim in our prediction. Another drop in performance can come from (ii) the size of the pretrained vector. PubMed vectors have smaller size of 200 compared to GloVe vectors which have a size of 300. Moreover, GloVe vector is trained in a larger corpus which makes the vector generalized more on our task. Previous research also showed that GloVe vectors can achieve higher performance in sentence similarity task which is more related to our task \cite{levy2015improving}. In the future, we will improve our understanding of the model by collecting and annotating more data from the other domains including Social Science and Computer Science. We will also explore the possibility of improving word vectors trained jointly on both scientific domain and large text corpus.

\subsection{Significance and future work}

Our automated claim extraction work contributes to the growing need to keep track of scientific knowledge being produced. As the number of publications grows significantly overtime, it has become impossible for any single researcher to keep track of scientific claims. Artificial Intelligence (AI) and Machine Learning (ML) will be needed for extracting these claims as they relate closely to many important aspects of knowledge production such as literature review and peer review \cite{weber2019applying}. For example, in literature review, readers want to find publications with similar claims or outcomes but that use different methods \cite{neves2019evaluation}. In peer review, reviewers want to have an up-to-date understanding of the same or related claims to estimate claim reliability of the submitted publication. A complete picture of the claims in the literature, obtained by methods such as the one proposed in our work, can therefore contribute to crucial components of the scientific process.

In this paper, we focus on claim extraction in the biomedical domain. This domain is by far the dominant producer of publications and therefore represents a good setting for automating claim extraction. Due to its diversity, biomedical research has differing concepts across its sub-fields. In our work, we show that we can take advantage of bigger datasets and pre-trained models of structured abstracts using transfer learning (Section 4). What is learned across sub-fields is effectively transferred to our specific dataset. This suggests that there is a common underlying set of concepts that can help other tasks. For example, some layers of our claim extraction network could be used in other common NLP tasks such as biomedical entity recognition and knowledge graph extraction. We will explore these avenues in the future.

What we have learned with this research suggests to us that we can further improve the model by increasing our dataset and by expanding it to other domains. For example, this would require us to scale our annotated biomedical corpus and to process and label other domains such as social science and economics. Our developed tool can be easily adjusted for these goals. Additionally, in the future, we would like to take advantage of recent developments in text encoding such as Bidirectional Encoder Representations from Transformers (BERT) \cite{devlin2018bert} or Generative Pre-Training (GPT) model \cite{radford2018improving}. Specific pre-trained models for science corpus have also been proposed which can be used in our task as well \cite{beltagy2019scibert}. These pre-trained models have shown to improve classification accuracy in various natural language processing tasks. We think that they can improve our predictions as well.

\section{Conclusion}

Automatically extracting claims made in scientific articles is becoming increasingly important for information retrieval task. In this paper, we present a novel annotated dataset for claim extraction from the biomedical domain and several claim extraction models drawing from recent developments in weak supervision and transfer learning. We show that transfer learning helps with the claim extraction task. We also release the data, the code, and a web service demo of our work so that the community can improve upon it. Claim extraction is an important task to automate, and our work helps in this direction.

Overall, with the substantial improvement we show here, we believe these models could be used in information extraction systems to support scholarly search engines such as \href{https://semanticscholar.org}{Semantic Scholar} or \href{https://www.ncbi.nlm.nih.gov/pubmed/}{PubMed}. By making the dataset and code available to the community, we hope to invite other researchers in the quest of analyzing scientific publications at scale.

\section{Acknowledgement}

This work was sponsored by Systematizing Confidence in Open Research and Evidence (SCORE) project from Defense Advanced Research Projects Agency (DARPA) and the Allen Institute for Artificial Intelligence (Allen AI). We thank our annotators for their great effort. Titipat Achakulvisut was partially supported by the Royal Thai Government Scholarship \#50AC002. Daniel E. Acuna was partially supported by \#1933803 and \#1800956.

\bibliography{references}

\begin{thebibliography}{10}

\bibitem{dernoncourt2017pubmed}
Franck Dernoncourt and Ji~Young Lee.
\newblock Pubmed 200k rct: a dataset for sequential sentence classification in
  medical abstracts.
\newblock {\em arXiv preprint arXiv:1710.06071}, 2017.

\bibitem{teufel2002summarizing}
Simone Teufel and Marc Moens.
\newblock Summarizing scientific articles: experiments with relevance and
  rhetorical status.
\newblock {\em Computational linguistics}, 28(4):409--445, 2002.

\bibitem{weber2019applying}
Rosina Weber.
\newblock Applying artificial intelligence in the science \& technology cycle.
\newblock {\em Information Services \& Use}, (Preprint):1--16, 2019.

\bibitem{webber2012discourse}
Bonnie Webber, Markus Egg, and Valia Kordoni.
\newblock Discourse structure and language technology.
\newblock {\em Natural Language Engineering}, 18(4):437--490, 2012.

\bibitem{liakata2010corpora}
Maria Liakata, Simone Teufel, Advaith Siddharthan, Colin~R Batchelor, et~al.
\newblock Corpora for the conceptualisation and zoning of scientific papers.
\newblock In {\em LREC}. Citeseer, 2010.

\bibitem{teufel2009towards}
Simone Teufel, Advaith Siddharthan, and Colin Batchelor.
\newblock Towards discipline-independent argumentative zoning: evidence from
  chemistry and computational linguistics.
\newblock In {\em Proceedings of the 2009 Conference on Empirical Methods in
  Natural Language Processing: Volume 3-Volume 3}, pages 1493--1502.
  Association for Computational Linguistics, 2009.

\bibitem{neves2019evaluation}
Mariana Neves, Daniel Butzke, and Barbara Grune.
\newblock Evaluation of scientific elements for text similarity in biomedical
  publications.
\newblock In {\em Proceedings of the 6th Workshop on Argument Mining}, pages
  124--135, 2019.

\bibitem{fisas2015discoursive}
Beatriz Fisas, Horacio Saggion, and Francesco Ronzano.
\newblock On the discoursive structure of computer graphics research papers.
\newblock In {\em Proceedings of The 9th Linguistic Annotation Workshop}, pages
  42--51, 2015.

\bibitem{dayrell2012rhetorical}
Carmen Dayrell, Arnaldo Candido~Jr, Gabriel Lima, Danilo Machado~Jr, Ann~A
  Copestake, Val{\'e}ria~Delisandra Feltrim, Stella~EO Tagnin, and Sandra~M
  Alu{\'\i}sio.
\newblock Rhetorical move detection in english abstracts: Multi-label sentence
  classifiers and their annotated corpora.
\newblock In {\em LREC}, pages 1604--1609, 2012.

\bibitem{chen2017improving}
Tao Chen, Ruifeng Xu, Yulan He, and Xuan Wang.
\newblock Improving sentiment analysis via sentence type classification using
  bilstm-crf and cnn.
\newblock {\em Expert Systems with Applications}, 72:221--230, 2017.

\bibitem{sateli2015semantic}
Bahar Sateli and Ren{\'e} Witte.
\newblock Semantic representation of scientific literature: bringing claims,
  contributions and named entities onto the linked open data cloud.
\newblock {\em PeerJ Computer Science}, 1:e37, 2015.

\bibitem{Habernal2014ArgumentationMO}
Ivan Habernal, Judith Eckle-Kohler, and Iryna Gurevych.
\newblock Argumentation mining on the web from information seeking perspective.
\newblock In {\em ArgNLP}, 2014.

\bibitem{Sardianos2015ArgumentEF}
Christos Sardianos, Ioannis~Manousos Katakis, Georgios Petasis, and Vangelis
  Karkaletsis.
\newblock Argument extraction from news.
\newblock In {\em ArgMining@HLT-NAACL}, 2015.

\bibitem{Dusmanu2017ArgumentMO}
Mihai Dusmanu, Elena Cabrio, and Serena Villata.
\newblock Argument mining on twitter: Arguments, facts and sources.
\newblock In {\em EMNLP}, 2017.

\bibitem{stab2014argumentation}
Christian Stab, Christian Kirschner, Judith Eckle-Kohler, and Iryna Gurevych.
\newblock Argumentation mining in persuasive essays and scientific articles
  from the discourse structure perspective.
\newblock In {\em ArgNLP}, pages 21--25, 2014.

\bibitem{freard2010role}
Dominique Fr{\'e}ard, Alexandre Denis, Fran{\c{c}}oise D{\'e}tienne, Michael
  Baker, Matthieu Quignard, and Flore Barcellini.
\newblock The role of argumentation in online epistemic communities: the
  anatomy of a conflict in wikipedia.
\newblock In {\em Proceedings of the 28th Annual European Conference on
  Cognitive Ergonomics}, pages 91--98. ACM, 2010.

\bibitem{Thorne2018FEVERAL}
James Thorne, Andreas Vlachos, Christos Christodoulopoulos, and Arpit Mittal.
\newblock Fever: a large-scale dataset for fact extraction and verification.
\newblock In {\em NAACL-HLT}, 2018.

\bibitem{yuan2019hclaime}
Shi Yuan and Bei Yu.
\newblock Hclaime: A tool for identifying health claims in health news
  headlines.
\newblock {\em Information Processing \& Management}, 56(4):1220--1233, 2019.

\bibitem{Guo2011AWA}
Yufan Guo, Anna Korhonen, and Thierry Poibeau.
\newblock A weakly-supervised approach to argumentative zoning of scientific
  documents.
\newblock In {\em EMNLP}, 2011.

\bibitem{dehghani2017learning}
Mostafa Dehghani, Aliaksei Severyn, Sascha Rothe, and Jaap Kamps.
\newblock Learning to learn from weak supervision by full supervision.
\newblock {\em arXiv preprint arXiv:1711.11383}, 2017.

\bibitem{Ratner2018TrainingCM}
Alexander Ratner, Braden Hancock, Jared Dunnmon, Frederic Sala, Shreyash
  Pandey, and Christopher R{\'e}.
\newblock Training complex models with multi-task weak supervision.
\newblock {\em CoRR}, abs/1810.02840, 2018.

\bibitem{augenstein2018multi}
Isabelle Augenstein, Sebastian Ruder, and Anders S{\o}gaard.
\newblock Multi-task learning of pairwise sequence classification tasks over
  disparate label spaces.
\newblock {\em arXiv preprint arXiv:1802.09913}, 2018.

\bibitem{dasigi2017experiment}
Pradeep Dasigi, Gully~APC Burns, Eduard Hovy, and Anita de~Waard.
\newblock Experiment segmentation in scientific discourse as clause-level
  structured prediction using recurrent neural networks.
\newblock {\em arXiv preprint arXiv:1702.05398}, 2017.

\bibitem{yu2019detecting}
Bei Yu, Yingya Li, and Jun Wang.
\newblock Detecting causal language use in science findings.
\newblock In {\em Proceedings of the 2019 Conference on Empirical Methods in
  Natural Language Processing and the 9th International Joint Conference on
  Natural Language Processing (EMNLP-IJCNLP)}, pages 4656--4666, 2019.

\bibitem{li2019causality}
Zhaoning Li, Qi~Li, Xiaotian Zou, and Jiangtao Ren.
\newblock Causality extraction based on self-attentive bilstm-crf with
  transferred embeddings.
\newblock {\em arXiv preprint arXiv:1904.07629}, 2019.

\bibitem{spacy2}
Matthew Honnibal and Ines Montani.
\newblock spacy 2: Natural language understanding with bloom embeddings,
  convolutional neural networks and incremental parsing.
\newblock {\em To appear}, 2017.

\bibitem{achakulvisut2015}
Achakulvisut Titipat and Daniel Acuna.
\newblock Pubmed parser: A python parser for pubmed open-access xml subset and
  medline xml dataset.
\newblock \url{http://github.com/titipata/pubmed_parser}, 2015.

\bibitem{Carletta1996AssessingAO}
Jean Carletta.
\newblock Assessing agreement on classification tasks: the kappa statistic.
\newblock {\em Computational Linguistics}, 22:249--254, 1996.

\bibitem{artstein2008inter}
Ron Artstein and Massimo Poesio.
\newblock Inter-coder agreement for computational linguistics.
\newblock {\em Computational Linguistics}, 34(4):555--596, 2008.

\bibitem{lauscher2018argument}
Anne Lauscher, Goran Glava{\v{s}}, and Simone~Paolo Ponzetto.
\newblock An argument-annotated corpus of scientific publications.
\newblock In {\em Proceedings of the 5th Workshop on Argument Mining}, pages
  40--46, 2018.

\bibitem{pennington2014glove}
Jeffrey Pennington, Richard Socher, and Christopher Manning.
\newblock Glove: Global vectors for word representation.
\newblock In {\em Proceedings of the 2014 conference on empirical methods in
  natural language processing (EMNLP)}, pages 1532--1543, 2014.

\bibitem{moen2013distributional}
SPFGH Moen and Tapio Salakoski2~Sophia Ananiadou.
\newblock Distributional semantics resources for biomedical text processing.
\newblock In {\em Proceedings of the 5th International Symposium on Languages
  in Biology and Medicine, Tokyo, Japan}, pages 39--43, 2013.

\bibitem{kingma2014adam}
Diederik~P Kingma and Jimmy Ba.
\newblock Adam: A method for stochastic optimization.
\newblock {\em arXiv preprint arXiv:1412.6980}, 2014.

\bibitem{gardner2018allennlp}
Matt Gardner, Joel Grus, Mark Neumann, Oyvind Tafjord, Pradeep Dasigi, Nelson
  Liu, Matthew Peters, Michael Schmitz, and Luke Zettlemoyer.
\newblock Allennlp: A deep semantic natural language processing platform.
\newblock {\em arXiv preprint arXiv:1803.07640}, 2018.

\bibitem{arora2016simple}
Sanjeev Arora, Yingyu Liang, and Tengyu Ma.
\newblock A simple but tough-to-beat baseline for sentence embeddings.
\newblock 2016.

\bibitem{levy2015improving}
Omer Levy, Yoav Goldberg, and Ido Dagan.
\newblock Improving distributional similarity with lessons learned from word
  embeddings.
\newblock {\em Transactions of the Association for Computational Linguistics},
  3:211--225, 2015.

\bibitem{devlin2018bert}
Jacob Devlin, Ming-Wei Chang, Kenton Lee, and Kristina Toutanova.
\newblock Bert: Pre-training of deep bidirectional transformers for language
  understanding.
\newblock {\em arXiv preprint arXiv:1810.04805}, 2018.

\bibitem{radford2018improving}
Alec Radford, Karthik Narasimhan, Tim Salimans, and Ilya Sutskever.
\newblock Improving language understanding with unsupervised learning.
\newblock Technical report, Technical report, OpenAI, 2018.

\bibitem{beltagy2019scibert}
Iz~Beltagy, Kyle Lo, and Arman Cohan.
\newblock Scibert: A pretrained language model for scientific text.
\newblock In {\em Proceedings of the 2019 Conference on Empirical Methods in
  Natural Language Processing and the 9th International Joint Conference on
  Natural Language Processing (EMNLP-IJCNLP)}, pages 3606--3611, 2019.

\end{thebibliography}
\bibliographystyle{unsrt}  


\end{document}